\documentclass[journal]{IEEEtran}

\usepackage{times}
\usepackage{epsfig}
\usepackage{graphicx}
\usepackage{amsmath}
\usepackage{amssymb}
\usepackage{tabulary}
\usepackage{multirow}
\usepackage{morefloats}
\usepackage{subfigure}
\usepackage{array}
\usepackage{soul}

\makeatletter
\newcommand{\removelatexerror}{\let\@latex@error\@gobble}
\makeatother

\makeatletter
\let\ftype@table\ftype@figure
\makeatother

\usepackage[pagebackref=true,breaklinks=true,letterpaper=true,colorlinks,bookmarks=false]{hyperref}

\begin{document}

\title{Planogram Compliance Checking Based on Detection of Recurring Patterns}




\author{Song Liu$^1$, Wanqing Li$^1$, Stephen Davis$^2$, Christian Ritz$^2$ and Hongda Tian$^1$\\
 $ $  \\
$^1$Advanced Multimedia Research Lab / School of Computing and Information Technology\\
$^2$Visual and Audio Signal Processing Lab / School of Electrical, Computer and Telecommunications Engineering \\
University of Wollongong, Wollongong, NSW, Australia, 2522\\
{\tt\small \{sl796, wanqing, stdavis, critz, ht615\}@uow.edu.au}}

\maketitle

\begin{abstract}
In this paper, a novel method for automatic planogram compliance checking in retail chains is proposed without requiring product template images for training. Product layout is extracted from an input image by means of unsupervised recurring pattern detection and matched via graph matching with the expected product layout specified by a planogram to measure the level of compliance. A divide and conquer strategy is employed to improve the speed. Specifically, the input image is divided into several regions based on the planogram. Recurring patterns are detected in each region respectively and then merged together to estimate the product layout. Experimental results on real data have verified the efficacy of the proposed method. Compared with a template-based method, higher accuracies are achieved by the proposed method over a wide range of products.
\end{abstract}



\section{Introduction}

\begin{figure}[ht]
\centering
\includegraphics[width=0.75\linewidth]{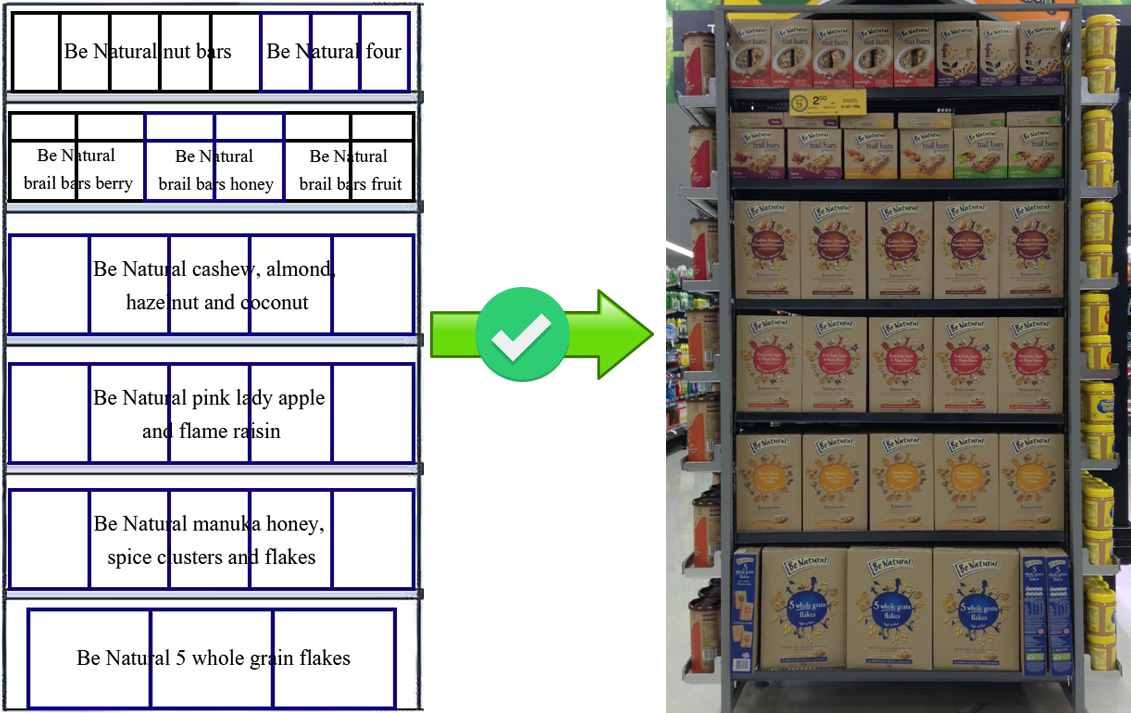}
\caption{An example of planogram compliance (left: a planogram specified by the company headquarters; right: a store shelf with a product layout that complies with the planogram).}
\label{fig:compliance_demo}
\end{figure}

\IEEEPARstart{I}{n} order to launch promotions and facilitate the best customer experience in retail chains, planograms are created to regulate how products should be placed on shelves. These planograms are usually created by the headquarters of a company and then distributed to its chain stores so that store managers can place products on shelves accordingly. Obviously, the company headquarters often need to verify whether each chain store would exactly follow the planograms. This verification process is referred to as planogram compliance checking. An example of a compliant product layout is shown in Fig \ref{fig:compliance_demo}. Conventional planogram compliance checking is conducted visually and manually, which is laborious and prone to human errors. As a result, many retail chains are seeking to automate this process recently.

Technologies based on computer vision have been explored for automatic planogram compliance checking.
In particular, the problem has been considered as a typical object detection problem in which products are detected and localized by matching input images of shelves against given template images of the products. Compliance checking can be then performed by comparing the detected product layout with the prespecified planogram. However, this approach requires up-to-date product template images, which are often unavailable. In addition, it is subject to the qualities of the images, light conditions, viewpoints of product template images and image pattern variations due to seasonal promotions which are carried out regularly by the product vendors.

For the purpose of promotion, it is noticed that multiple instances of a product are usually required to be displayed consistently on a shelf.  These multiple instances within the input image form similar yet non-identical visual objects and can be referred to as a \textbf{recurring visual pattern} or \textbf{recurring pattern}~\cite{6619105}. By detecting the recurring pattern, the product instances that form the pattern can be localised.  It is expected that the layout of the shelf could be estimated by detecting all the recurring patterns and locating all the instances in each patterns. Then the estimated layout could be compared with the expected product layout specified by a planogram to measure the level of compliance. Since detection of the recurring patterns does not require template images for training, the compliance checking does not require any template images of the products on the shelf.

This paper proposes a novel method for automatic planogram compliance checking without requiring product template images. The method mainly consists of (1) estimation of product layout from the results of recurring pattern detection; (2) compliance checking by comparing the estimated product layout with a planogram using spectral graph matching~\cite{Leordeanu:2005:STC:1097115.1097816}; and (3) a divide and conquer strategy to improve the speed. This paper is an extension of the work presented in \cite{liu_compliance}. The extension includes more justification of the proposed algorithm, comparison to a template-based method and refinement of the compliance checking using the product images automatically extracted from input images through recurring pattern detection.

The remainder of the paper is organized as follows. Section II reviews some representative methods for automatic planogram compliance checking and recurring pattern detection. The details of the proposed method are presented in Section III. Experimental results are provided and discussed in Section IV. Finally, the paper is concluded with some perspectives on future work in Section V.

\section{Related Work}

\subsection{Automatic Planogram Compliance Checking}

Conventional automatic methods for planogram compliance checking involve extracting product layout information based on well-established object detection and recognition algorithms, which usually require template images as training samples. Auclair et al. \cite{Auclair:2007:USV:1422772.1422797} presented a product detection system from input images by matching with existing templates using scale-invariant feature transform (SIFT) vectors. A real-time online product detection tool using speeded up robust features (SURF) and optical flow was proposed in \cite{5543576}, which also depends on high-quality training data. Another study focused on product logo detection by spatial pyramid mining \cite{4607625}.
A recent method carried out by Varol and Kuzu \cite{doi:10.1117/12.2179127} utilised a cascade object detection framework and support vector machine (SVM) to detect and recognise cigarette packages on shelves, which also requires template images for training. Despite these progresses in estimating product layout information by means of object detection and recognition, most methods require either strong or weak supervision for object modeling. Although some unsupervised approaches based on latent topic models \cite{Sivic05discoveringobject, Karlinsky:2008:UCP:1479250.1479277} have been proposed, they still need images for learning. 

\subsection{Recurring Pattern Detection}

Multiple instances or objects of the same product on a shelf share similar visual appearance. In particular, these objects sharing similar groups of visual words can be formulated as recurring patterns. By detecting those recurring patterns using an unsupervised object-level matching method, product layout could be extracted without requiring template images of the products on the shelf.
In the literature, detecting recurring patterns is referred to as common visual pattern discovery \cite{4408869, 5539780}, co-recognition/segmentation of objects\cite{5539777, 1640859, Cho:2008:CIP:1478237.1478249}, and high-order structural semantics learning \cite{5459465}. There are three typical approaches for recurring pattern detection:
\begin{itemize}
\item pairwise visual word matching which matches pairs of visual words across all objects \cite{5459465};
\item pairwise visual object matching which matches feature point correspondences between a pair of objects \cite{5539780, 5539777}; and
\item pairwise visual word-object matching which matches visual words and objects simultaneously \cite{6619105}.
\end{itemize}
In \cite{1640859, 4269998} unsupervised detection/segmentation of two objects in two images was explored. Yuan and Wu \cite{4408869} detected object pairs from one single image or an image pair using spatial random partitioning. Cho et al. \cite{Cho:2008:CIP:1478237.1478249} achieved the same goal by solving a correspondence association problem via markov chain monte carlo (MCMC) exploration. As for pairwise object matching based methods for detecting multiple recurring patterns, Liu and Yan \cite{5539780} employed graph matching to detect recurring patterns between two images. Agglomerative clustering \cite{5459322} and MCMC association \cite{5539777} were adopted by Cho et al. to deal with multiple object matching. Gao et al. \cite{5459465} used a pairwise visual word matching approach to detect recurring patterns. Liu and Liu \cite{6619105} discovered recurring patterns from one image by optimizing a pairwise visual word-object joint assignment problem using greedy randomized adaptive search procedure (GRASP) \cite{Feo95greedyrandomized}.

Both visual words and objects are considered in pairwise visual word-object joint assignment,
which could yield results with higher detection accuracy than methods
where only visual word matching or object matching is performed.
As a result, pairwise visual word-object matching \cite{6619105} is adopted for recurring pattern detection in the proposed method.
However, solving such a joint assignment problem is computational expensive especially in detecting recurring patterns with a large number of visual objects. To improve the speed, a divide and conquer strategy is proposed to partition the image into regions to control the number of visual objects in each region.

\section{Proposed Method}

\begin{figure*}
\begin{center}
\includegraphics[width=6.8in]{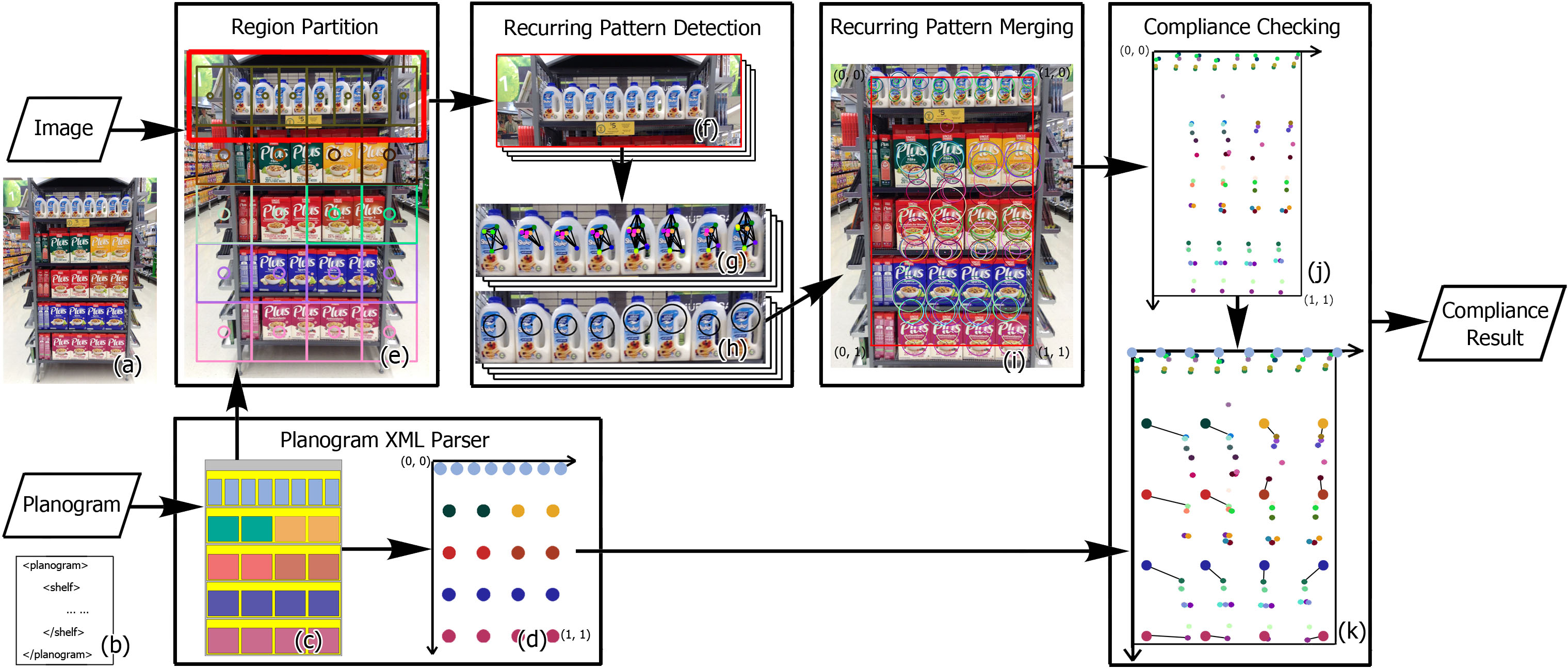}
\end{center}
   \caption{The block diagram of the proposed method for planogram compliance checking.
   (a) input image;
   (b) the corresponding planogram in XML;
   (c) parsed planogram shown in 2D boxes;
   (d) 2D points representing expected product layout ($PointSet_{planogram}$);
   (e) product boxes and an estimated region projected to the input image;
   (f) a region for one type of product;
   (g) a detected recurring pattern shown in visual words and objects;
   (h) a detected recurring pattern shown in circular regions;
   (i) bounding box with merged recurring patterns;
   (j) 2D points representing detected product layout ($PointSet_{detected}$);
   (k) searching for the optimal matches using graph matching and greedy algorithm.}
\label{compliance_flow}
\end{figure*}

In the proposed method, an input image is firstly partitioned into regions based on the information parsed from a planogram. Repeated products are detected in each region and then merged together to estimate product layout. Finally, the estimated product layout is compared against the expected product layout specified in the planogram for compliance checking. The block diagram of the proposed method is shown in Fig \ref{compliance_flow}, the details of which will be provided in the following sections.

\
\subsection{Planogram XML Parser and Region Partition}

A planogram is created by company headquarters to indicate how and where specific products should be placed on shelves. Therefore, the layout information stored in a planogram can be regarded as the expected layout of the corresponding input image. Moreover, such layout information can be used to divide the input image into regions corresponding to different types of products.

In the proposed method, an input planogram is stored in XML format, which needs to be parsed to retrieve related formation for region partition and compliance checking. A parsed planogram for a particular shelf contains the following information:
\begin{itemize}
\item number of rows in this shelf,
\item number of products in each row, and
\item type of each product.
\end{itemize}
A planogram is illustrated in Fig \ref{compliance_flow}(c) where
the shelf, every row in the shelf and every product could be represented as 2D boxes.
Due to the shape of a regular retail shelf, a rectangle (grey box in Fig \ref{compliance_flow}(c)) is used to represent the whole shelf.
Then the shelf is vertically divided into several identical rows (yellow boxes in Fig \ref{compliance_flow}(c)) according to the number of rows in this shelf.
Finally, each row is horizontally divided into boxes according to the number of products that are placed in that row. As a result, products can be represented by boxes in each row.

Given the estimated boxes for each product, product position and layout are described using a set of 2D points.
A 2D coordinate system is created by considering the top-left corner of the shelf box as the origin $(0, 0)$ and bottom-right corner as $(1, 1)$. The expected product layout can be then represented by all the centre points of product boxes in this coordinate system which are denoted as $PointSet_{planogram}$ (Fig \ref{compliance_flow}(d)):
\begin{equation}
PointSet_{planogram}=\{P_1,...,P_M\};
\label{psp}
\end{equation}
\begin{equation}
P_i=\{p_{i1},p_{i2}, ..., p_{im}\},
\label{pi}
\end{equation}
where $P_i$ is the set of points that correspond to the $i^{th}$ type of product specified in the planogram
(i. e. points in the same colour in Fig \ref{compliance_flow}(d)).
$p_{ii'}$ is a point with the 2D coordinate $(x_{p_{ii'}}, y_{p_{ii'}})$, where $x_{p_{ii'}}, y_{p_{ii'}} \in [0,1]$, $|PointSet_{planogram}|=M$.

In order to estimate regions, all product boxes are overlapped to the input image.
For each type of product, the regions covered by the product boxes of that type are grouped into a rectangular search region.
The search region is then extended by a margin to allow for differences in the product locations on the shelf compared to the planogram specification.
An example of this area is labelled as the red box in Fig \ref{compliance_flow}(e).

Considering the speed of recurring pattern detection and the known minimum size of products listed in the planogram that are to be displayed, every region will be limited to include no more than 25 product instances.  If there are a large number of instances of the same type of product within a region, the region for this product type will be further divided to meet the criteria above. The choice of the maximum number of product instances in a search region is to balance the time spent on the recurring pattern detection and merge of the detected patterns. It was empirically chosen to be $25$ in this paper.

\subsection{Recurring Pattern Detection and Merging}

The recurring pattern detection aims to estimate a product layout by finding and locating near-identical visual objects from one single input image.
After region partition, the original input image is divided into several regions, at least one for each type of product.  Therefore, recurring patterns are detected in each region and later merged together for layout estimation.

The method presented in \cite{6619105} is adopted which realizes recurring pattern discovery by solving a simultaneous visual word-object assignment problem. In the method, a recurring pattern is defined as a 2D feature-assignment matrix where each row corresponds to a visual word and each column corresponds to a visual object. Detected feature points are populated in the matrix to ensure feature points in each row/visual word (i.e. points in the same colour in Fig \ref{compliance_flow}(g)) share strong appearance similarity while layouts formed by points in every column/visual object (i.e. connected black lines in Fig \ref{compliance_flow}(g)) share strong geometric similarity. An energy function is defined to achieve both appearance and geometric consistencies.

This joint assignment problem is NP-hard and is thus optimized by a greedy randomized adaptive search procedure (GRASP) with matrix operations called local moves that are specifically designed for convergence purpose. In these local moves, operations are taken to expand and maintain the 2D feature-assignment matrix which include adding/deleting a row/column and modifying entries. In each iteration of GRASP, local moves are applied stochastically to explore a variety of local optima. Each optimized assignment matrix can be regarded as a detected recurring pattern.

As a result, for each input region, a set of multiple candidate recurring patterns could be detected. Circular regions are calculated to represent a recurring pattern where each detected visual object is covered by a circle (Fig \ref{compliance_flow}(h)). Therefore, detected recurring patterns can be written as follows:
\begin{equation}
CandidatePatterns=\{Pattern_1, ..., Pattern_n\};
\label{pattern_1}
\end{equation}
\begin{equation}
Pattern_i=\{(x_{i1},y_{i1},r_{i1}),...,(x_{in},y_{in},r_{in})\}.
\label{pattern_2}
\end{equation}
In a recurring pattern $Pattern_i$, each circular region is represented by a centre with a 2D coordinate and a radius. The centre is calculated as the mean position of feature points of the visual object. The radius is the mean value of width and height of the bounding box covering all feature points taken by the visual object.
Assuming $Pattern_s$ and $Pattern_t$ are detected from different regions that actually belong to the same product type, whether $Pattern_s$ and $Pattern_t$ will be merged depends on the coverage of their circular regions. If a circular region from $Pattern_s$ overlaps with another circular region from $Pattern_t$, these two regions will be combined into one.


Recent improvement of the method presented in \cite{6619105} allows extracting image patches for all detected visual objects from the input image. These patches can serve as product images to refine compliance checking results. More details are presented in Section IV.C.

\subsection{Compliance Checking}
To match with the expected product layout $PointSet_{planogram}$ from the planogram,
another group of point sets representing detected product layout $PointSet_{detected}$ (Fig \ref{compliance_flow}(j)) needs to be constructed by processing the detected recurring patterns.

First, all the visual object centres of each pattern are regarded as 2D points. Then a minimum bounding box is calculated to cover all points. A 2D coordinate system is created by considering the top-left corner of this bounding box as the origin (0, 0) and bottom-right corner as the coordinate (1, 1). 2D coordinates are calculated for all the detected visual object centres in such a coordinate system (Fig \ref{compliance_flow}(i)). The object centre points from the $j^{th}$ recurring pattern in this 2D coordinate system are denoted as point set $R_j$.
$PointSet_{detected}$ can be constructed as follows:
\begin{equation}
PointSet_{detected}=\{R_1,...,R_N\};
\label{rsr}
\end{equation}
\begin{equation}
R_j=\{r_{j1},r_{j2}, ..., r_{jn}\},
\label{ri}
\end{equation}
where $r_{jj'}$ is a point with the 2D coordinate $(x_{r_{jj'}}, y_{r_{jj'}})$; $x_{r_{jj'}}, y_{r_{jj'}} \in [0,1]$; $|PointSet_{detected}|=N$, the number of detected products. The height and width of bounding box are denoted as $Height_{box}$ and $Width_{box}$ respectively.

\subsubsection{Graph Matching}

Given the point set $P$ from $PointSet_{planogram}$, containing $m$ points, and $R$ from $PointSet_{detected}$, containing $n$ points,
a matching between the points in $P$ and those in $R$ is performed by solving a spectral graph matching problem\cite{Leordeanu:2005:STC:1097115.1097816}.

The affinity matrix $U$ for graph matching is created as follows by considering geometric layout relations between any pair of assignments $(a, b)$ where $a = (p_i, r_{i'})$ and $b = (p_j, r_{j'})$; $p_i, p_j \in P$; $r_{i'}, r_{j'} \in R$.
\begin{equation}
U(a,b)=\exp(-\frac{\Delta_{dh}^2}{\delta_{dh}}-\frac{\Delta_{dv}^2}{\delta_{dv}});
\end{equation}
\begin{equation}
\Delta_{dh}=\max(dis\_h(p_i ,r_{i'} ), dis\_h(p_j,r_{j'} ));
\end{equation}
\begin{equation}
\Delta_{dv}=\max(dis\_v(p_i ,r_{i'} ), dis\_v(p_j,r_{j'} )),
\end{equation}
where $\delta_{dh}$ and $\delta_{dv}$ are weight parameters.
In the experiments, $\delta_{dh}=(Width_{box}/(Height_{box}+Width_{box}))^2$ and $\delta_{dv}=(Height_{box}/(Height_{box}+Width_{box}))^2$. $dis\_h()$ and $dis\_v()$ return the horizontal and vertical distance respectively.
$U \in R^{k \times k}$ is a sparse symmetric and positive matrix where $k = m \times n$.

The matching problem is to find a cluster $C$  of assignments $(p_i, r_{i'})$ which could maximize the score $S=\Sigma_{a,b \in C} U (a, b)$ regarding the undirected weighted graph that is represented by $U$. This cluster $C$ can be described by an indicator vector $x$, such that $x(a) = 1$ if $a \in C$ and $x(a) = 0$ otherwise. The total inter-cluster score can be rewritten as:

\begin{equation}
S=\sum_{a, b \in C} U(a, b)=x^TUx.
\end{equation}
The optimal solution $x^*$ is the binary vector that could maximize the score:
\begin{equation}
x^*=argmax(x^TUx).
\end{equation}
$x^*$ that will maximize the score $x^T U x$ is the principal eigenvector of $U$. A greedy algorithm is then applied to the principal eigenvector of $U$ making it the binary indicator vector $x$. The matching score $S=\Sigma_{a,b \in C} U (a, b)=x^TUx$ is calculated and stored for compliance checking.

\subsubsection{Check Compliance}
$M*N$ possible matching cases can be generated by spectral graph matching.
The matching scores of these cases form an $M$ by $N$ matrix that indicates matching confidences between any expected product layout $P_i$ from $PointSet_{planogram}$ and detected product layout $R_j$ from $PointSet_{detected}$.
Therefore, a greedy algorithm is performed again on this matrix to identify the optimal matches from $PointSet_{planogram}$ to $PointSet_{detected}$ (Fig \ref{compliance_flow}(k)).
The greedy algorithm firstly accepts the match with the maximum matching score and then rejects all other matches that are conflicting with the accepted one.
The process repeats until all scores in the matching matrix are either accepted or rejected.
After the greedy matching, every type of product specified in the planogram will finally be matched with one unique and optimal recurring pattern.

Each product from these optimal recurring patterns will be marked as compliant (true positive) if it can be matched with one product from the planogram. Detected product that cannot be matched with any products from the planogram will be marked as being not compliant (false positive). Moreover, if a product from the planogram cannot be matched with any products from recurring patterns, the position of this product will be marked as empty which is another instance of being not compliant.


\section{Experimental Results}

To perform an experimental validation of the proposed method, a data set was collected from a supermarket chain, where all images were captured using an iPad Mini 1st Gen (5MP camera) due to the fact that images from popular mobile devices are easy to obtain.
Additionally, the proposed method is implemented in C++ on a PC with 3.4GHz Intel\textsuperscript{R} Core\textsuperscript{TM} i7 CPU.


The testing cases in the real data set are of different product size, product quantity and feature quality.
Feature quality could directly affect the accuracy of compliance checking. Products with poor feature quality will lead to insufficient visual words in finding recurring patterns, which would in turn result in false detection.
Product quantity refers to the number of instances that belong to the same type of product.
This number can greatly affect the speed of compliance checking, as the processing time of recurring pattern detection will increase dramatically with the increase of product quantity.

Based on these characteristics of testing cases, experiments were conducted to evaluate both the effectiveness and the speed of the proposed method. Due to the limit of space, results on some selected samples with different characteristics (shown in Table \ref{testing_cases}) are presented in this paper.


\begin{table}[!t]
\renewcommand{\arraystretch}{1.3}
\caption{Characteristics of selected samples}
\label{testing_cases}
\centering
\begin{tabular}{|c|c|c|c|}
\hline
{\bf \begin{tabular}[c]{@{}c@{}}Selected\\ Sample\end{tabular}} & {\bf \begin{tabular}[c]{@{}c@{}}Product\\ Size\end{tabular}} & {\bf \begin{tabular}[c]{@{}c@{}}Product\\ Quantity\end{tabular}} & {\bf \begin{tabular}[c]{@{}c@{}}Feature\\ Quality\end{tabular}} \\ \hline
Toilet Paper                                                    & Big                                                          & Small                                                            & Rich                                                            \\ \hline
Heater                                                          & Big/Medium                                                   & Medium/Small                                                     & Medium                                                          \\ \hline
Coke                                                            & Big                                                          & Medium/Small                                                     & Rich                                                            \\ \hline
Cereal                                                          & Medium                                                       & Small                                                            & Rich                                                            \\ \hline
Shampoo                                                         & Small                                                        & Medium/Small                                                     & Poor                                                            \\ \hline
Tissue                                                          & Small                                                        & Large                                                            & Medium                                                          \\ \hline
Chocolate                                                       & Small                                                        & Large                                                            & Poor                                                            \\ \hline
\end{tabular}
\end{table}

\subsection{Evaluation on Effectiveness}


For each testing case, an image and its corresponding planogram XML file served as the inputs.
For each product type, a number of matched products $N_{matched}$ from graph matching and a number of expected products from the planogram $N_{expected}$ are compared. The compliance accuracy for one product type is calculated as $1-|N_{matched}-N_{expected}|/N_{expected}$. The compliance accuracy of a testing case is calculated by averaging compliance accuracies over all product types contained in the case.
For testing purpose, planogram XML files are created manually to match the product layout in the input images. Therefore, the compliance accuracy for each testing case could also indicate the accuracy of the algorithm.

For comparison purpose, a template-based method for planogram compliance checking was implemented as a baseline. The baseline follows a conventional idea of using product template images for training and detection. Specifically, for each input image, the algorithm needs to be trained using corresponding product template images which consists of detecting SIFT keypoints and extracting SIFT descriptors. Then the input image is divided into several non-overlapping segments and a brute-force matching is carried out on each segment until all product templates are exhausted.
A match is found if the number of matched descriptors between the input and template images exceeds a threshold.
Product template images in our experiments are cropped from the images taken by a high-resolution camera.

The overall accuracy achieved by the proposed method is 90.53\% whereas the accuracy by the template-based method is 71.84\%.
Accuracies of both methods with respect to product size, quantity and feature quality are provided in Table
\ref{accuracy_size_comp}, Table \ref{accuracy_quantities_comp} and Table \ref{accuracy_quality_comp} respectively.
The experimental results have shown that the proposed method is effective for planogram compliance checking. Compared with the template-based method, the proposed method achieves higher compliance checking accuracies especially when dealing with products that are of small size, large quantity, or poor feature quality. Although the accuracy of the template-based method could be improved for some cases of small products or large quantity by using the planogram as a prior, it could not achieve the same accuracy as the proposed one does. As expected, the accuracies dropped for the testing cases with small product size, large product quantity and poor feature quality. Obviously, lower feature quality will lead to worse compliance checking accuracy. As for the small product size, the main reason is that smaller products captured in images tend to possess limited texture features. Moreover, smaller sized products tend to be packed in a large quantity on the shelf, which contributes to the decreased accuracy when dealing with products with large quantities.

In general, the false positive rates are very low for both methods, and false positives tend to happen for the same category of products when their packages look very similar, but they are classified as different products, for instance, different tissues and shampoos from the same vendors. On average, the false positives were less than $1\%$ for the template-based method and less than $3\%$ for the proposed method over the entire set of testing images.
 
Besides, the proposed method is also effective in dealing with non-front-view testing cases, whereas the template-based method is not suitable due to the fact that most product templates are front-view images. The variations brought by different view points will further degrade its accuracy.

\begin{table}[!t]
\renewcommand{\arraystretch}{1.3}
\caption{Compared compliance checking accuracies for groups with different characteristics.}
\label{table:accuracy}
\centering
\subtable[Accuracies for group with different product sizes]{
\label{accuracy_size_comp}
\begin{tabular}{|c|c|c|}
\hline
\multirow{2}{*}{\textbf{\begin{tabular}[c]{@{}c@{}}Product\\ Size\end{tabular}}} & \multicolumn{2}{c|}{\textbf{Compliance Accuracy}}                                                    \\ \cline{2-3}
                                                                                 & \multicolumn{1}{l|}{\textbf{Template-Based  Method }} & \multicolumn{1}{l|}{\textbf{Proposed Method }} \\ \hline
Big                                                                              & 94.60\%                                              & 95.90\%                                       \\ \hline
Medium                                                                           & 89.82\%                                              & 90.61\%                                       \\ \hline
Small                                                                            & 40.34\%                                             & 84.44\%                                       \\ \hline
\end{tabular}
}
\subtable[Accuracies for group with different product quantities]{
\label{accuracy_quantities_comp}
\begin{tabular}{|c|c|c|}
\hline
\multirow{2}{*}{\textbf{\begin{tabular}[c]{@{}c@{}}Product\\ Quantity\end{tabular}}} & \multicolumn{2}{c|}{\textbf{Compliance Accuracy}}                                                    \\ \cline{2-3}
                                                                                     & \multicolumn{1}{l|}{\textbf{Template-Based Method }} & \multicolumn{1}{l|}{\textbf{Proposed Method }} \\ \hline
Large                                                                                & 51.69\%                                              & 87.57\%                                       \\ \hline
Medium                                                                               & 79.43\%                                              & 87.85\%                                       \\ \hline
Small                                                                                & 91.25\%                                              & 92.94\%                                       \\ \hline
\end{tabular}
}
\subtable[Accuracies for group with different feature qualities]{
\label{accuracy_quality_comp}
\begin{tabular}{|c|c|c|}
\hline
\multirow{2}{*}{\textbf{\begin{tabular}[c]{@{}c@{}}Feature\\ Quality\end{tabular}}} & \multicolumn{2}{c|}{\textbf{Compliance Accuracy}}                                                    \\ \cline{2-3}
                                                                                    & \multicolumn{1}{l|}{\textbf{Template-Based Method }} & \multicolumn{1}{l|}{\textbf{Proposed Method }} \\ \hline
Rich                                                                                & 95.71\%                                              & 96.03\%                                       \\ \hline
Medium                                                                              & 82.49\%                                              & 91.77\%                                       \\ \hline
Poor                                                                                & 55.95\%                                              & 81.24\%                                       \\ \hline
\end{tabular}
}
\end{table}

\begin{figure}[!htbp]
\centering
\subfigure[][]{
\includegraphics[width=.65\linewidth]{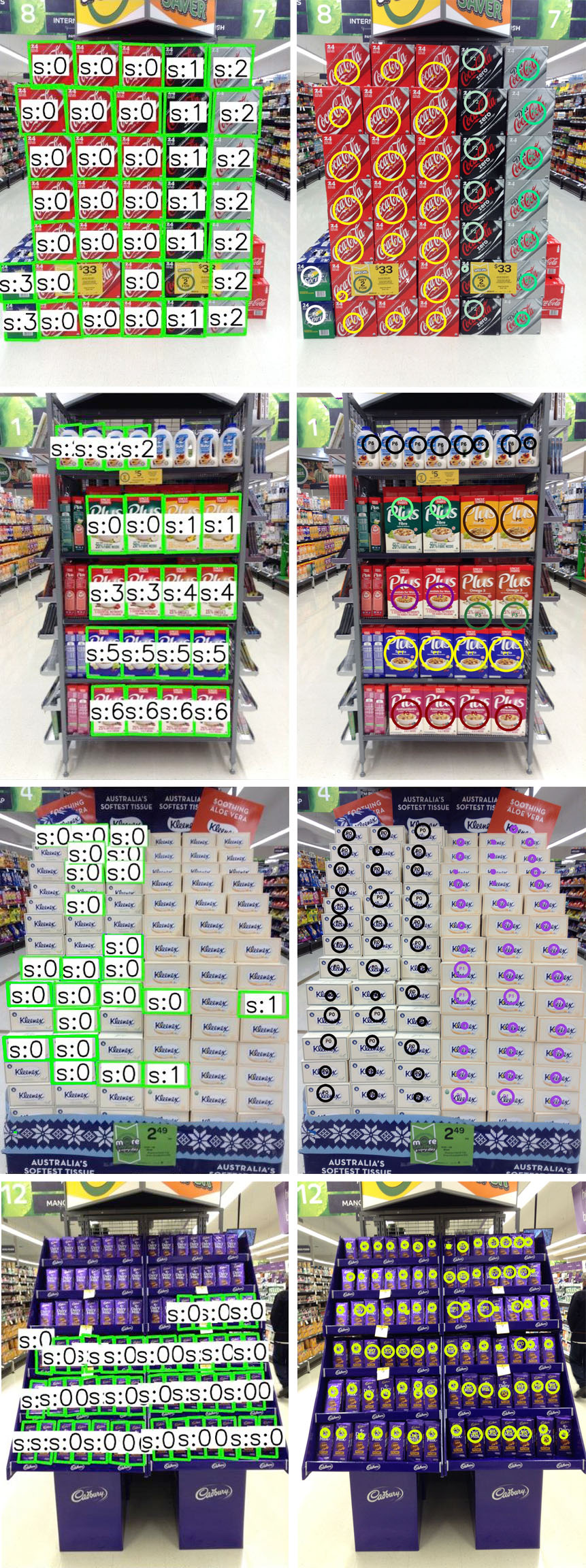}
\label{fig:results_comp}}
\subfigure[][]{
\includegraphics[width=1\linewidth]{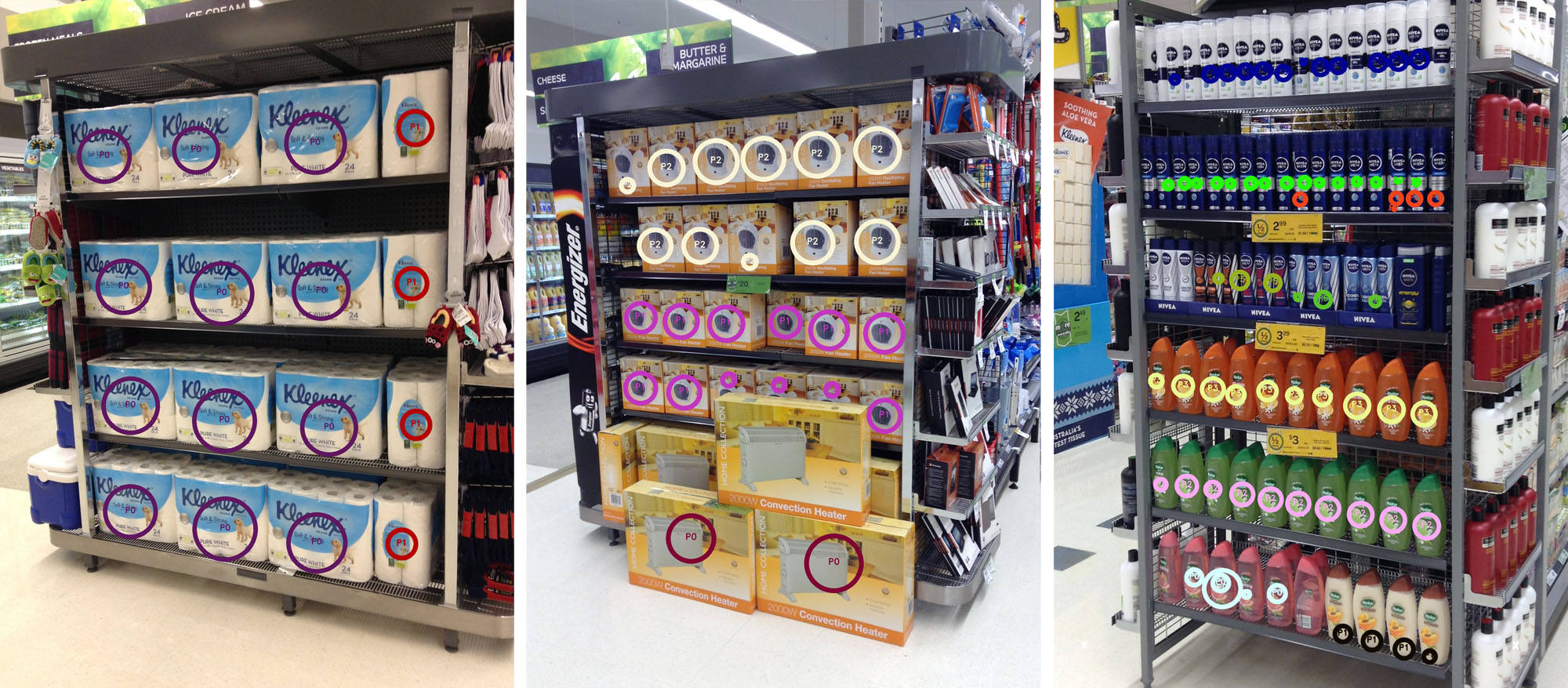}
\label{fig:results_v}}
\caption{
Detected products on some selected samples.
\subref{fig:results_comp} compared results on selected front-view samples from both methods (left: the temple-based method; right: the proposed method; samples include coke, cereal, tissue and chocolate);
\subref{fig:results_v} results of proposed method on some non-front-view samples including toilet paper, heater and shampoo.
}
\label{fig:results_comgp_v}
\end{figure}

Results on some selected front-view samples are shown in Fig \ref{fig:results_comp}, where detected products are marked in input images. For the template-based method, detected products are highlighted using green bounding boxes with labels indicating different product types detected.
For the proposed method, detected products of the same type are labelled using circles with the same colour.
Additionally, results of the proposed method on some selected non-front-view samples are presented in Fig \ref{fig:results_v}.
For intermediate results on recurring pattern merging, please refer to our work presented in \cite{liu_compliance}.

\subsection{Evaluation on Speed}

The speed of the proposed method is assessed by using the average time that is required to process one testing case.
Results on the speed for both methods are illustrated in Table \ref{time_quantity_comp}.
In general, the proposed method will cost more time than the template-based method especially in the cases with a large product quantity.

For the proposed method, most computation is actually spent on the detection of recurring patterns. The computational complexity of the recurring pattern detection algorithm is $O(n^3)$, where $n$ is the number of visual objects which is the product quantity in our case. Therefore, The adopted divide and conquer strategy that partitions the image into regions to control the product quantity in each region can effectively reduce the overall CPU time especially in the cases involving a large number of products.

To validate the improvement on speed brought by the divide and conquer method, further experiments were carried out on the cases with large product quantities. In these experiments, the number of regions was adjusted and the processing time v.s. the  number of regions is shown in Table \ref{ptime}. Empirically, it has shown that region partition could averagely improve the speed by over 70\% without compromising the compliance checking accuracy.

\begin{table}[!t]
\renewcommand{\arraystretch}{1.3}
\caption{Experimental results on speed.}
\label{table:efficiency}
\centering
\subtable[Average processing time from both methods for group with different product quantities]{
\begin{tabular}{|c|c|c|}
\hline
\multirow{2}{*}{\textbf{\begin{tabular}[c]{@{}c@{}}Product\\ Quantity\end{tabular}}} & \multicolumn{2}{c|}{\textbf{Average Processing Time}}                                                \\ \cline{2-3}
                                                                                     & \multicolumn{1}{l|}{\textbf{Template-Based  Method }} & \multicolumn{1}{l|}{\textbf{Proposed Method }} \\ \hline
Large                                                                                & 274 seconds                                          & 478 seconds                                   \\ \hline
Medium                                                                               & 166 seconds                                          & 359 seconds                                   \\ \hline
Small                                                                                & 147 seconds                                          & 127 seconds                                   \\ \hline
\end{tabular}
    \label{time_quantity_comp}
}
\subtable[Processing time of the proposed method using different number of regions]{
\begin{tabulary}{1.0\textwidth}{|c|c|c|c|c|}
\hline
\multicolumn{3}{|c|}{\textbf{Tissue Sample: 39 products of the same type}}                                                                                                                                                                                         \\ \hline
\begin{tabular}[c]{@{}c@{}}\textbf{Number of} \\ \textbf{Regions}\end{tabular} & \begin{tabular}[c]{@{}c@{}} \textbf{Expected Number of} \\ \textbf{Products in Each Region}\end{tabular} & \begin{tabular}[c]{@{}c@{}}\textbf{Processing} \\ \textbf{Time}\end{tabular} \\ \hline
1                                                                & 39                                                                                   & 489 seconds                                                                      \\ \hline
2                                                                & 20                                                                                   & 195 seconds                                                  \\ \hline
3                                                                & 13                                                                                   & 136 seconds                                                  \\ \hline
\end{tabulary}
    \label{ptime}
}
\end{table}

\subsection{Evaluation on Product Image Extraction}

\begin{figure}[ht]
\centering
\subfigure[][]{
\includegraphics[width=0.6\linewidth]{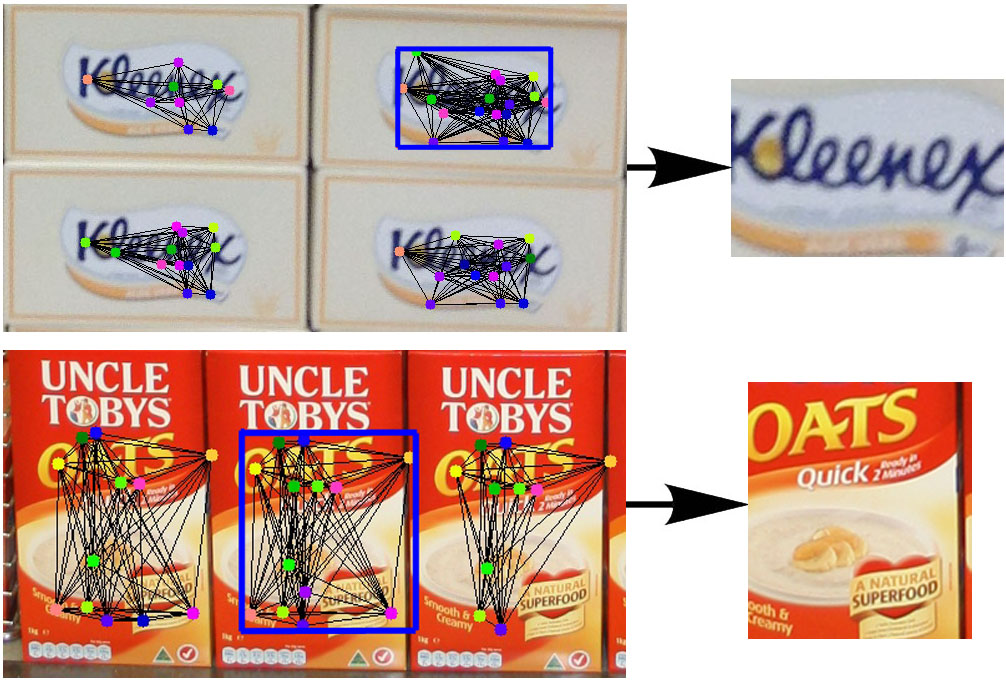}
\label{fig:patch}}
\subfigure[][]{
\includegraphics[width=1\linewidth]{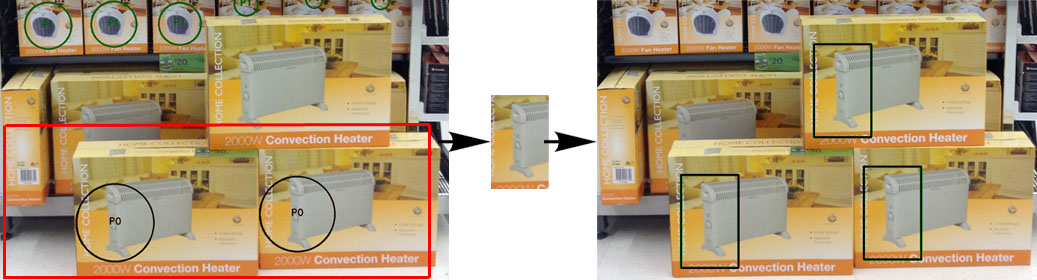}
\label{fig:temp_det}}
\caption{
Results on product image extraction and compliance checking refinement.
\subref{fig:patch} examples of generating product image from recurring patterns;
\subref{fig:temp_det} examples of using the extracted product image to re-detect products (the red box
indicates the region of recurring pattern detection).}
\label{fig:patch temp}
\end{figure}

The proposed method is also capable of extracting product images. Based on the graph matching results, each type of product can be linked to a unique recurring pattern. In this recurring pattern, each visual object can be regarded as a product instance and can be represented by a set of feature points (Fig \ref{compliance_flow}(g)). A bounding box which covers all these feature points could well represent this product (see Fig \ref{fig:patch}). In order to find out the most suitable rectangular region that could represent a particular product type, the product instance in a detected recurring pattern that possesses the most feature points is considered as the most representative one. The rectangular region of this product instance is then selected as the product image of its product type. Example results on selecting product images from recurring patterns are shown in Fig \ref{fig:patch}.

The extracted product images could be useful in many ways. For instance, it can be used to detect products that were missed during recurring pattern detection. This would further improve the compliance checking accuracy. Some products may not be detected due to an unsuitable region partition. However, missing products may still be found by matching extracted product images within the whole input image. One example is shown in Fig 6, where a heater on the top of the other two is not detected as the region partition fails to include all heaters. The missing heater is picked up by using the extracted product image and a template matching algorithm.

\section{Conclusion and Future Work}

To the best of our knowledge, there is no existing automatic method for planogram compliance checking without using template images. A novel approach for extracting retail shelf layouts using recurring pattern detection is thus presented in this paper. The method detects products effectively and efficiently by merging detected recurring patterns from divided regions of the input image. Given product layout information, planogram compliance checking and product matching are performed by solving a graph matching problem.
The proposed method can also extract product images automatically which can be used to further improve the compliance checking result.
Compared with a template-based method, the proposed method is much more effective especially when processing low quality images.
The proposed method is challenged by deformable packages such as chips. This will be our future focus.

\section*{Acknowledgement}

This work is partially supported by Smart Services CRC Australia.

{\small
\bibliographystyle{ieee}
\bibliography{compliance}
}

\end{document}